\documentclass{article}

\usepackage[preprint]{cpal_2024}

\usepackage{hyperref}
\usepackage{url}
\usepackage{resizegather}
\usepackage{amsmath}%
\usepackage{MnSymbol}%
\usepackage{wasysym}%
\usepackage{graphicx}
\usepackage{rotating}
\usepackage{array,makecell,multirow}
\usepackage{resizegather}
\usepackage{tikz}
\newcommand*\circled[1]{\tikz[baseline=(char.base)]{
            \node[shape=circle,draw,inner sep=0.4pt] (char) {#1};}}
\definecolor{CColor}{rgb}{0.01,0.31,0.59}
\usepackage[export]{adjustbox}
\usepackage{xcolor,colortbl}
	
\definecolor{LightCyan}{rgb}{0.88,1,1}
\usepackage{tabularx}
\definecolor{Gray}{gray}{0.90}
\newcolumntype{g}{>{\columncolor{Gray}}c}
\definecolor{ffe1da}{RGB}{255,225,218}
\definecolor{F7E0D5}{RGB}{247,224,213}
\definecolor{darkF7E0D5}{RGB}{209,154,128}
\usepackage{wrapfig}
\usepackage{subcaption}

\usepackage{url}

\title{The Counterattack of CNNs in Self-Supervised Learning:  Larger Kernel Size might be All You Need}

\author{%
  Tianjin Huang\textsuperscript{1},Tianlong Chen\textsuperscript{2}, Zhangyang Wang\textsuperscript{3},  ~Shiwei Liu\textsuperscript{1,4}\\
  \textsuperscript{1}Eindhoven University of Technology,  \textsuperscript{2} Massachusetts Institute of Technology\\ \textsuperscript{3}The University of Texas at Austin,  \textsuperscript{4}University of Oxford\\
  \texttt{t.huang@tue.nl, s.liu3@tue.nl, tianlong@mit.edu, atlaswang@utexas.edu }
}

\begin{document}

\maketitle

\begin{abstract}
Vision Transformers have been rapidly uprising in computer vision thanks to their outstanding scaling trends, and gradually replacing convolutional neural networks (CNNs). Recent works on self-supervised learning (SSL) introduce siamese pre-training tasks, on which Transformer backbones continue to demonstrate ever stronger results than CNNs. People come to believe that Transformers or self-attention modules are inherently more suitable than CNNs in the context of SSL. However, it is noteworthy that most if not all prior arts of SSL with CNNs chose the standard ResNets as their backbones, whose architecture effectiveness is known to already lag behind advanced Vision Transformers. Therefore, it remains unclear whether the self-attention operation is crucial for the recent advances in SSL - or CNNs can deliver the same excellence with more advanced designs, too? Can we close the SSL performance gap between Transformers and CNNs? To answer these intriguing questions, we apply self-supervised pre-training to the recently proposed, stronger lager-kernel CNN architecture and conduct an apple-to-apple comparison with Transformers, in their SSL performance. Our results show that we are able to build pure CNN SSL architectures that perform on par with or better than the best SSL-trained Transformers, by just scaling up convolutional kernel sizes besides other small tweaks. Impressively, when transferring to the downstream tasks \texttt{MS COCO} detection and segmentation, our SSL pre-trained CNN model (trained in 100 epochs) achieves the same good performance as the 300-epoch pre-trained Transformer counterpart. We hope this work can help to better understand what is essential (or not) for self-supervised learning backbones. Codes will be made public.
\end{abstract}

\section{Introduction}

After leading the computer vision field for a couple of decades, the dominant position of convolutional neural networks (CNNs)~\citep{Alexnet,vgg,resnet,densenet,howard2017mobilenets,xie2017aggregated,efficientnet} is being vigorously challenged by the emerging Vision Transformers~\citep{dosovitskiy2021an,touvron2021training,wang2021pyramid,vaswani2021scaling,yuan2021tokens,zhai2022scaling,d2021convit, liu2021swin}. The emergence of local-window  self-attention~\citep{liu2021swin,liu2021swin2,vaswani2021scaling,yang2021focal} significantly unleashes the power of Transformers as general backbones taking over various computer vision benchmarks rapidly with permissible resource budget, including ImageNet classification~\citep{dosovitskiy2021an}, region-level object detection~\citep{dong2022cswin}, dense pixel-level semantic segmentation~\citep{zheng2021rethinking}, and video action classification~\citep{arnab2021vivit}.  

Recently, the prominent representation power of Transformer with self-attention is also introduced into the self-supervised learning (SSL) regime. Previous art~\citep{trinh2019selfie} mimics the masked language modeling and conducts a preliminary exploration on masked patch prediction using ResNet~\citep{resnet}. iGPT~\citep{chen2020generative} pre-trains sequence Transformers to predict pixels in an auto-regressive way as a generative model, and adopted a linear probing for classification. Dosovitskiy et al.~\citep{dosovitskiy2021an} further pre-train vanilla ViT on large-scale JFT-300M dataset, showing the promise of ViT on self-supervision. MoCo-v3~\citep{chen2021empirical} generalizes contrastive learning to ViT achieving 84.1\% accuracy on ImageNet-1K~\citep{russakovsky2015imagenet}. DINO~\citep{caron2021emerging} proposes a self-distillation self-supervision paradigm where two ViT models fed with the same image but different views are trained to minimize their output probability distribution. They show that under this form of self-supervision, ViT explicitly contains segmentation information. Recently, EsViT~\citep{li2021efficient} and MoBY~\citep{xie2021self} illustrate that more advanced Swin Transformer~\citep{liu2021swin} can also be applied with SSL, standing out as a better backbone than ViT and CNNs. However, it is worth noting that most if not all previous arts chose the standard ResNets as the CNN backbone whose architectural design is known to already lag behind advanced Vision Transformers~\citep{liu2022convnet}, rendering unfair comparisons between Transformers and CNNs.

Since the backbone choice is a crucial ingredient to SSL, it is more desirable to draw a relatively fair comparison between CNNs and Transformers to better understand if the self-attention in Transformers is crucial to the recent advances in SSL, or if CNNs with state-of-the-art designs can have the same promise.  In this paper, we turn to the recently proposed, stronger CNN architecture - ConvNeXt~\citep{liu2022convnet}. By modernizing the seemingly ``old-fashioned'' ResNet towards the design of Swin Transformer, ConvNeXt favorably rivals Swin Transformer on ImageNet classification and downstream tasks, which can therefore conduct a more apple-to-apple comparison with Transformers in the context of SSL. Our work is intended to test whether the recent strike of CNNs can be generalized to the SSL regime, as well as build a new state-of-the-art baseline for self-supervised CNNs in the era of Transformers. We briefly summarize our contributions below:

\begin{itemize}
    \item [$\blacksquare$] An intriguing phenomenon is first observed in our paper: while ConvNeXt demonstrates compelling performance over strong Swin Transformers in supervised learning, its performance in SSL is no better than the original Transformer backbone - ViT. 
    \item [$\blacksquare$] Nevertheless, simply adding two small adaptions (i. naively scaling the kernel size up; ii. adding Batchnorm layers after depthwise convolutions) to vanilla ConvNeXt, we are able to build attention-free CNNs, which we dubbed Big ConvNet SSL (\textbf{BC-SSL}), to perform on par or even better than the best SSL-trained Transformers with linear probe and $k$-NN evaluation on ImageNet classification,  while enjoying faster inference throughput (up to 40\% faster on A100 GPU).
    
    \item [$\blacksquare$] More impressively, when transferring to downstream tasks such as linear classification, detection, and segmentation, our modified CNN architecture demonstrates significantly larger performance gains. Simply as it is, our SSL pre-trained BC-SSL (trained in 100 epochs) achieves equally good performance to the 300-epoch pre-trained Swin Transformer counterparts on \texttt{MS COCO} object detection and segmentation.
    
     \item [$\blacksquare$] We also observe an encouraging trend of robustness evaluation, that is, the robustness of BC-SSL monotonously improves as the kernel size scales up to 15$\times$15, performing an all-around win over Swin-T in terms of both clean and robust accuracy.
    
\end{itemize}
We mainly focus on probing self-supervised large-kernel CNNs using ConvNext in this work, yet we are aware of other CNN architectures positively equipped with even larger kernels like RepLKNet~\citep{ding2022scaling} and SLaK~\citep{liu2022more}, which could also be competitive baselines in this regime. Although we observe that the benefits of large kernels seem to saturate at 9$\times$9 kernels in self-supervised ConvNeXt, we do not exclude the possibility that other architectures such as RepLKNet or SLaK can benefit more from increasing kernels further, which we leave as future work.

\section{Related Work}
\subsection{Visual Self-Supervised Learning}
Most if not all self-supervised learning methods in computer vision can be categorized as discriminative or generative~\citep{grill2020bootstrap}.

Contrastive learning is a leading direction in  discriminative approaches that achieve state-of-the-art SSL performance~\citep{SimCLR,he2020momentum,CPC,CPCv2,DIM,AMDIM,he2020momentum,chen2020improved,chen2021empirical}. Contrastive methods avoid a costly pixel-level generation step and aim to learn augmentation-invariant representation by bringing the representation between different augmented pairs of the same image (positive pairs) closer and pushing the representation of augmented views from different images (negative pairs) away from each other~\citep{wu2018unsupervised,doersch2017multi,SimCLR}. A drawback of this approach is
the requirement of comparing features from a large number
of images (including positive pairs and negative pairs) simultaneously. More importantly, such an approach usually needs a large batch of data~\citep{SimCLR} or memory banks~\citep{he2020momentum,wu2018unsupervised} to obtain sufficient negative pairs. 

Many works start to propose various techniques to eliminate the negative pairs due to the cumbersome comparisons between different examples. DeepCluster~\citep{caron2018deep} successfully avoids the usage of negative pairs by applying a clustering process. More specifically, it uses the  representation from the prior phase to cluster data points, after which the cluster index of the data point is treated as the classification target for the new representation. 
Follow-up work continues to improve the effectiveness and efficiency of  simultaneous clustering and representation learning~\citep{asano2019self,caron2018deep,caron2019unsupervised,huang2019unsupervised,li2020prototypical,zhuang2019local}. BYOL~\citep{grill2020bootstrap} is another milestone work that effectively removes the negative pairs with strong results. BYOL feeds two networks with different augmented views of the same image. The online network is trained online to predict the representation of the target network whose weights are updated with a slow-moving average (momentum encoder) of the former. The momentum encoder was claimed to be crucial to prevent collapse.  A follow-up study~\citep{chen2021exploring} shows that stop-gradient operation plays an essential role in preventing collapsing and BYOL works even without a momentum encoder at some performance cost. Inspired by  mean teacher~\citep{tarvainen2017mean} and BYOL, DINO uses a self-distillation-based loss instead of a contrastive loss achieving strong SSL performance with ViT~\citep{dosovitskiy2021an}. EsViT~\citep{li2021efficient} recently explore DINO to Swin Transformers. They propose to match the region-level features together with the view-level features for multi-stage Transformers and further establish a new state-of-the-art bar for SSL.  

Generative methods seek to jointly learn data and representation together~\citep{donahue2016adversarial,donahue2019large,brock2018large,donahue2016adversarial} with either auto-encoding of images~\citep{vincent2008extracting,kingma2013auto,rezende2014stochastic} or adversarial learning~\citep{goodfellow2020generative}.  Recent
generative approaches revisit the mask language modeling in images as pre-training tasks have achieved competitive
finetuning performance~\citep{dosovitskiy2021an,bao2021beit,he2022masked,zhou2021ibot,xie2022simmim}. 

The vast majority of recent breakthroughs achieved in SSL are accompanied by advanced Transformer architecture. Therefore, it is important to decouple SSL from Transformers and to see whether self-attention-free architectures like CNN can deliver the same excellence with more advanced designs too.

\begin{wrapfigure}{r}{0.55\textwidth}
\begin{center}
\includegraphics[width=0.45\textwidth]{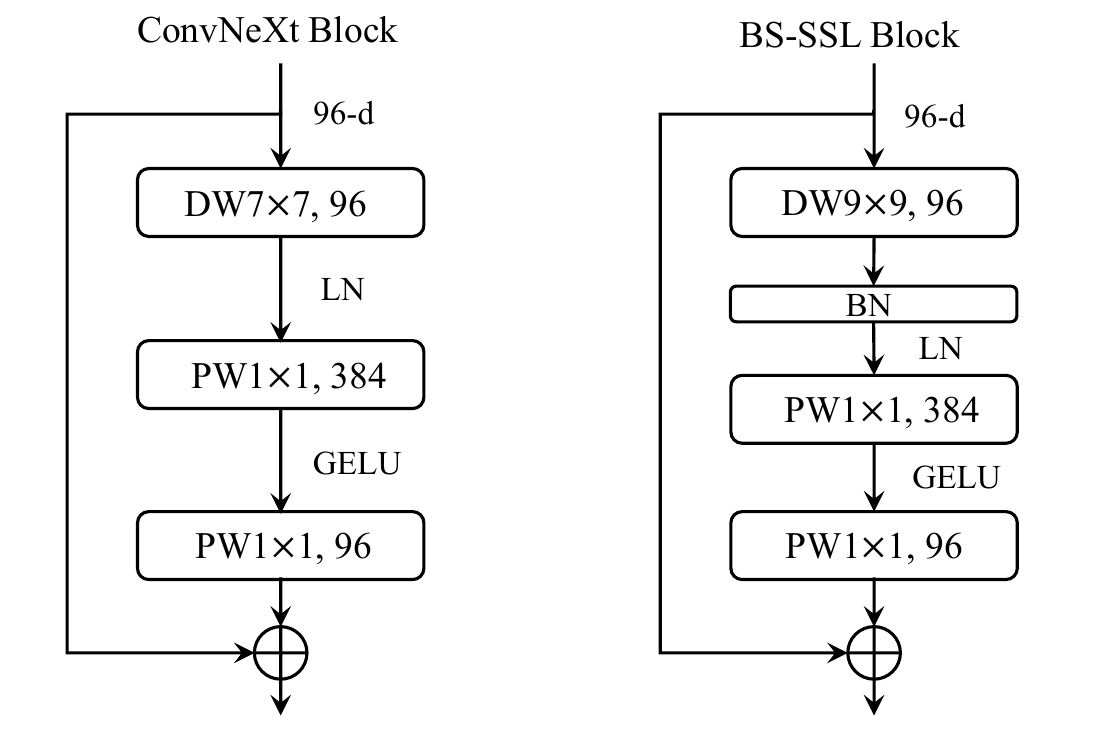}
\caption{\textbf{Block designs for ConvNeXt and BC-SSL.} ``DW'' refers to depthwise convolution and ``PW'' stands for pointwise convolution. To fully translate the promise of modern CNNs in supervised learning to self-supervised learning, two small adaptions are adopted on the original ConvNeXt: (1) adding BatchNorm layers after large depthwise kernels; (2) naively scaling up convolutional kernel size to 9$\times$9.}
\label{fig:BC-SSL}
\end{center}
\end{wrapfigure}

\subsection{Large Kernels in Supervised Learning}
Large kernels in supervised learning have a long history, stemming from the 2010s~\citep{krizhevsky2012imagenet,szegedy2015going,szegedy2017inception}, where AlexNet~\citep{krizhevsky2012imagenet} adopts 11$\times$11 kernels in the first convolutional layer for instance. Global Convolutional Network~\citep{peng2017large} replaces a 2D convolution with two parallels of stacked 1D convolution, with kernel size up to 25$\times$1 $+$ 1$\times$25. The idea has recently been revisited in SegNeXt~\citep{guo2022segnext} to build an efficient multi-scale attention module. Perhaps predominantly due to the popularity of VGG~\citep{simonyan2014very}, people start to blindly stack multiple small kernels (i.e., 1$\times$1 or 3$\times$3) to obtain a large receptive field for computer vision tasks~\citep{resnet,howard2017mobilenets,xie2017aggregated,densenet}.

Motivated by the large window size of Swin Transformer, ConvNeXt~\citep{liu2022convnet} explores the inverted bottleneck design equipped with 7$\times$7 kernels, evincing the promise of large kernels holds for CNN. RepLKNet~\citep{ding2022scaling} is a concurrent work that scales kernel size to 31$\times$31 using an auxiliary 5$\times$5 kernel. SLaK~\citep{liu2022more} pushes the kernel size to 51$\times$51 by employing certain decomposition and sparsity techniques, improving the training stability and memory scalability of large convolutions kernels.  
More recently,~\citet{chen2022scaling,xiao2022dynamic} reveals the feasibility of large kernels for 3D CNNs and time series classification too, respectively. However, the potential of modern large-kernel CNNs has never been explored in the context of self-supervised learning. Our paper conducts a pilot study asking whether we can close the SSL performance gap between Transformers and CNNs by introducing these stronger large-kernel CNN architectures.

\begin{wrapfigure}{r}{0.4\textwidth}
\vspace{-2em}
\begin{center}
\includegraphics[width=0.4\textwidth]{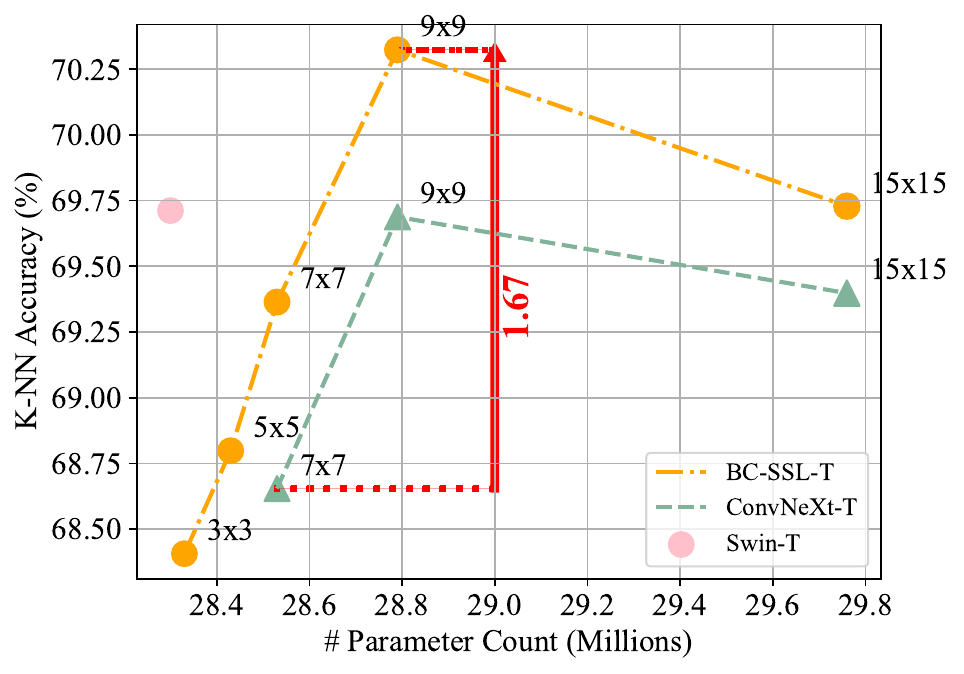}
\caption{\textbf{$k$-NN Accuracy of ConvNeXt-T, BC-SSL-T, and Swin-T with various kernel sizes.} BC-SSL-T improves the $k$-NN accuracy over the standard ConvNeXt-T by 1.67\%. Models are pre-trained for 100 epochs with DINO.}
\label{fig:small_scale}
\end{center}
\vspace{-3.5 em}
\end{wrapfigure}

\section{Modern Large-Kernel CNNs in Self-Supervised Learning}
In this section, we will carry out the exploration of the modern large-kernel CNNs in self-supervised learning. A brief recap of modern large-kernel CNNs is provided first, followed by the evaluation of the vanilla ConvNeXt in SSL showing inferior performance to its Transformers competitors. We consequently study several design choices built upon which we can bridge the performance gap between modern large-kernel CNNs between SoTA Transformers, i.e., Swin Transformers, in SSL. 

\subsection{A Brief Recap of ConvNeXt}
ConvNeXt~\citep{liu2022convnet}, a recently emerging pure CNN model armed with 
more sophisticated architecture design, heats up the debate between CNNs and Transformers in supervised learning. It thoroughly investigates the architecture designs used in Swin Transformers and assembles a set of principles that substantially boosts the performance of a standard ResNet-50 to the level of the state-of-the-art Transformers.  Specifically, ConvNeXt adopts a different stage compute ratio (1:1:9:1); a ViT-style patchify stem that divides images into non-overlapped patches; depthwise convolution and inverted bottleneck with an increase of network width; 7$\times$7 large kernels instead of 3$\times$3; replacing ReLU~\citep{nair2010rectified} with GELU~\citep{hendrycks2016gaussian}; substituting BatchNorm~\citep{ioffe2017batch} with LayerNorm~\citep{ba2016layer}. Built upon the following set of principles, ConvNeXt is able to match or even outperform Swin Transformers under most scenarios in supervised learning. Ding et al.~\citep{ding2022scaling} and Liu et al.~\citep{liu2022more} further push along the direction of large kernel and demonstrate the possibility of scaling kernel size up to 31$\times$31 and 51$\times$51, while achieving even better performance, respectively.

\subsection{Vanilla ConvNeXt in SSL}

		
\begin{wraptable}{r}{0.6\textwidth}
	\caption{\textbf{Linear and $k$-NN classification on ImageNet-1K.} Models are pre-trained for 100 epochs with DINO.}
	\label{tab:vanilla_convnext}
	\centering
	\begin{tabular}{l|cccc}
		\hline
		Model & Kernel Size & Param (M) & Linear & $k$-NN \\
		\hline
		ResNet-50 & 3$\times$3 & 23 & 67.14 & 58.88  \\
		\hline
		ConvNeXt-T & 3$\times$3 & 28 & 73.27 & 67.98  \\
		\hline
		ConvNeXt-T & 5$\times$5 & 28 & 73.82  & 68.34  \\
		\hline
		ConvNeXt-T & 7$\times$7 & 29 & 74.10  & 68.65 \\
		\hline
		ViT-S &  - & 21 & 73.51 & 68.77 \\
		\hline
		Swin-T & -  &28 & \textbf{74.98} & \textbf{69.72} \\
		\hline
	\end{tabular}
	\vspace{-0.1in}
\end{wraptable}

We first examine whether the vanilla ConvNeXt can close the performance gap between Transformers and CNNs. We choose the state-of-the-art self-distillation with no labels (DINO)~\citep{caron2021emerging} as our SSL framework with 4 architectures: ResNet-50, ConvNeXt-T, ViT-S with 16$\times$16 patch size, and Swin-T with 4$\times$4 patch size. We choose ConvNeXt-T as the representative of modern CNN architectures mainly due to its several similarities with the SoTA SSL-trained Swin-T~\citep{liu2021swin} in terms of (1) parameter count; (2) throughput; (3) and supervised performance. We follow the vanilla \underline{100-epoch} training configuration used in DINO and pre-train models on ImageNet-1K without labels with AdamW~\citep{loshchilov2018decoupled} and a batch size of 512. Cosine learning rate decay is used whose base learning rate is scaled with the batch size as $lr = 0.0005*\text{batchsize} / 256$. Weight decay is also decayed from 0.04 to 0.4 with a cosine function. The teacher temperature of self-distillation $\tau$ is set as 0.4. We report the results of two evaluation protocols $k$-NN and linear probing in Table~\ref{tab:vanilla_convnext}. 

We observe that even with the standard $3\times3$ kernels, the superb architecture design of ConvNeXt directly brings 6.13\% higher linear probing accuracy over the ``old-fashion'' ResNet. Increasing kernel size from 3$\times$3 to 7$\times$7 continuously boosts the accuracy by 0.83\%. However, the performance of standard ConvNeXt with 7$\times$7 kernels still falls behind its Transformer competitors. The above observation indicates that \textit{the promise of modern CNN designs in supervised learning seems can not be fully translated to the SSL scenario.}

\subsection{Pushing the Limits of ConvNeXt in SSL}

\paragraph{Adding BatchNorm after large depthwise kernels.} By default, ConvNeXt is a BatchNorm-free architecture. Substituting Batchnorm with LayerNorm slightly improves the performance as reported in the original work. Albeit that BatchNorm may have many intricacies that cause detrimental effects on performance~\citep{wu2021rethinking} in supervised learning, it historically plays an essential role in self-supervised learning. It has been shown that batch normalization is crucial for BYOL to achieve good performance~\citep{grill2020bootstrap,richemond2020byol}.  MoCo-v3~\citep{chen2021empirical} shows that removing all BatchNorm layers in the MLP heads causes a 2.1\% accuracy drop. MoBY~\citep{xie2021self} confirms a similar phenomenon in Swin Transformers.

In contrast to these previous arts, we investigate the effect of batch normalization in the backbone. More specifically, we choose to add a BatchNorm layer after each depthwise convolutional kernel in the ConvNeXt backbone. Table~\ref{tab:bn} demonstrates that this small modification consistently improves the performance of self-supervised ConvNeXt across various batch sizes. 

\begin{wraptable}{r}{0.6\textwidth}
\small
\caption{\textbf{Linear classification of ConvNeXt with vs. without BatchNorm (BN) after Depthwise convolutions on ImageNet-1K.} Models are pre-trained for 100 epochs with DINO.
\label{tab:bn}}
	\begin{center}
		
		\begin{tabular}{l|ccccc}
			\hline
		Model &  \multicolumn{5}{c}{Linear Classification} \\
        \hline
        Kernel Size & 3$\times$3 & 5$\times$5 & 7$\times$7 &  9$\times$9 & 15$\times$15  \\
        \hline
        ConvNeXt-T w/o BN & 73.27 & 73.82 & 74.10 & 74.52 & 74.28  \\
        ConvNeXt-T w/ BN & \textbf{74.04} & \textbf{74.35} &  \textbf{74.48} & \textbf{75.01} & \textbf{74.47} \\

		\hline
		\end{tabular}
        \vspace{-0.1 in}
	\end{center}
\centering
\end{wraptable}

\paragraph{Naively scaling up convolutional kernel sizes.} The larger kernel benefit of ConvNeXt reaches a saturation point at 7$\times$7 in supervised learning~\citep{liu2022convnet}. While it is possible to expand performance gains by further enlarging kernels, sophisticated techniques like structure re-parameterization~\citep{ding2022scaling} and sparsity~\citep{liu2022more} are required. Nonetheless, we empirically find that in self-supervised learning the kernel size can be favorably scaled to 9$\times$9 without bells and whistles as shown in Table~\ref{tab:bn}. However, an accuracy drop occurs when we further increase the kernel size to 15$\times$15. Although increasing kernel sizes beyond 9$\times$9 does not provide more performance gains for ConvNeXt, we do not exclude the possibility that other large-kernel recipes such as RepLKNet and SLaK can benefit more from increasing kernel further, as we have observed much improvement room of promise in supervised learning~\citep{ding2022scaling,liu2022more}. Potential future work is how to fully explore the capacity of larger kernels beyond 9$\times$9 in self-supervised learning.

Until now, we have finished our exploration of modern large-kernel CNNs in self-supervised learning and ended up with our modified architecture in Figure~\ref{fig:BC-SSL}, which we dubbed \textbf{BC-SSL}. The above two small tweaks (adding BatchNorm after depthwise convolutions and scaling convolutional kernel sizes to 9$\times$9) bring encouraging performance gains. Figure~\ref{fig:small_scale} shows that the added BatchNorm consistently brings around 0.7\% $k$-NN accuracy gains to ConvNeXt across kernel sizes, and enlarging kernels from 7$\times$7 to 9$\times$9 further increases the performance by 0.97\%, outperforming Swin Transformers. Overall we achieve an encouraging 1.67\% accuracy improvement over the vanilla ConvNeXt even under a small-scale pre-training regime. 

In the next section, we will evaluate the scalability of BC-SSL in terms of training time and model size, as well as the transferability on downstream tasks. \textit{From now on, we will choose 9$\times$9 as our default kernel size and add a BatchNorm layer after depthwise convolutions in each residual block for BC-SSL.}

		

\section{Main Evaluations of BC-SSL} 
\looseness=-1  We first evaluate the proposed BC-SSL backbone in SSL with the standard self-supervised benchmark on ImageNet-1K~\citep{russakovsky2015imagenet}. We also evaluate the quality of the learned representations by conducting downstream transfer learning on MS COCO detection and segmentation~\citep{lin2014microsoft}, 18 small datasets, and several ImageNet-level robustness benchmarks. 

\subsection{Evaluation on ImageNet-1K} 

\begin{table*}[h]
\small 
\centering
\caption{Comparison with SoTA SSL results across different architectures on ImageNet-1K. The patch size is 16$\times$16 and 4$\times$4 for ViT and Swin Transformers, respectively, and the window size of Swin Transformers is set as default 7$\times$7. ViT-BN is ViT that replaces the LayerNorm before MLP blocks by BatchNorm.  ``RN152w2+SK'' refers to ResNet-152 with 2$\times$ wider channels and selective kernels~\citep{li2019selective}. Throughput numbers are obtained from DINO~\citep{caron2021emerging} except for Swin Transformer and BC-SSL, which are measured by us using a V100 GPU (black) and an A100 GPU (\textcolor{blue}{blue}), respectively, following~\citep{liu2022convnet}. On an A100 GPU, SSL pre-trained BC-SSL can have a much higher (up to 40\%) throughput than SSL pre-trained Swin Transformer.}
\label{tab:main_result_cls}
  \resizebox{0.9\textwidth}{!}{
\begin{tabular}{lccccc}
\hline
 Method & Architecture  &  \#Parameters (M) $\downarrow$~~ &  Throughput $\uparrow$ & Linear $\uparrow$~~ & $k$-NN $\uparrow$  \\ 
\hline
\rowcolor{Gray}
\multicolumn{6}{c}{\em SoTA SSL with Big Model Sizes}  \\
iGPT~\citep{chen2020generative} & iGPT-XL &  6801 & - & 72.0 & - \\

SCLR~\citep{SimCLR}   & RN50w4 & 375 & 117 & 76.8 & 69.3 \\
SwAV~\citep{caron2020unsupervised} & RN50w5 & 586 & 76 & 78.5 &  67.1 \\
BYOL~\citep{grill2020bootstrap} & RN50w4 & 375 & 117 & 78.6 & -- \\
MoCo-v3~\citep{chen2021empirical} & ViT-H-BN/16 & 632 & 32 & 79.1 & - \\
SimCLR-v2~\citep{chen2020big} & RN152w2+SK  & 354 & - & 79.4 & - \\
BYOL~\citep{grill2020bootstrap}   & RN200w2  & 250 & 123 & 79.6 & 73.9 \\
\hline
\rowcolor{Gray}
\multicolumn{6}{c}{\em SoTA SSL with Small Model Sizes}  \\
BYOL~\citep{grill2020bootstrap}   & RN50  & 23 & 1237 & 74.4 & 64.8 \\
MoBY~\citep{xie2021self} &  Swin-T & 28 &  758/\textcolor{blue}{\scriptsize 1326} & 75.1 & - \\
DCv2~\citep{caron2020unsupervised}   & RN50  & 23 & 1237 & 75.2 & 67.1 \\
SwAV~\citep{caron2020unsupervised} & RN50 & 23 & 1237 &  75.3 & 65.7 \\
DINO~\citep{caron2021emerging} &  RN50 & 23 &  1237 &  75.3   & 67.5 \\
MoCo-v3~\citep{chen2021empirical} &  ViT-B &  85  & 312 & 76.7 & - \\
DINO~\citep{caron2021emerging} & ViT-S & 21 & 1007 & 77.0 & 74.5 
\\
DINO~\citep{li2021efficient} &  Swin-T & 28 &  758/\textcolor{blue}{\scriptsize 1326} &  77.0   & 74.2 \\
\rowcolor{LightCyan}
DINO (Ours) &  BC-SSL-T & 29 &  762/\textcolor{blue}{\scriptsize 1777 (+34\%)} &  77.8   & 75.7 \\
DINO~\citep{caron2021emerging} & ViT-B & 85 & 312 & 78.2 & 76.1  \\

DINO~\citep{li2021efficient} & Swin-S & 49 &  437/\textcolor{blue}{\scriptsize 857} &  79.2  & 76.8 \\

\rowcolor{LightCyan}
DINO (Ours) & BC-SSL-S & 50 &  442/\textcolor{blue}{\scriptsize 1197 (+40\%)} &  79.0  & 76.6 \\

\hline
\end{tabular} }
\end{table*}

\begin{table*}[h]
\tiny
\centering
\caption{\textbf{Object detection and segmentation on MS COCO.} Models are pre-trained on ImageNet-1K and finetuned using Mask-RCNN for 36 epochs. The performance of BC-SSL is implemented by us and the results of other models are obtained from their original papers.}
  \resizebox{0.9\textwidth}{!}{
\begin{tabular}{l|c|c|cccccc}
\hline
Method & Architecture & Kernel Size & AP$^{box}$ $\uparrow$  & AP$^{box}_{50}$ $\uparrow$   & AP$^{box}_{75}$ $\uparrow$  & AP$^{mask}$ $\uparrow$  & AP$^{mask}_{50}$ $\uparrow$   & AP$^{mask}_{75}$ $\uparrow$  \\
\hline
  \rowcolor{Gray}
  \multicolumn{9}{c}{\em 300-epoch supervised learning}  \\
Supervised~\citep{liu2022convnet} & Swin-T & -  & 46.0   &  68.1  &  50.3   &  41.6   &  65.1  &  44.9 \\
Supervised~\citep{liu2022convnet}  & ConvNeXt-T & 7$\times$7   & 46.2   &  67.9  &  50.8   &  41.7   &  65.0  &  44.9 \\
\hline
  \rowcolor{Gray}
  \multicolumn{9}{c}{\em 100-epoch SSL pre-training}  \\
\rowcolor{LightCyan}
DINO & BC-SSL-T & 3$\times$3  & 45.0 & 66.3 & 49.6 & 40.5 & 63.5 & 43.6 \\
\rowcolor{LightCyan}
DINO & BC-SSL-T & 5$\times$5  &45.9 & 67.3 & 50.6 & 41.1 & 64.3 & 44.3 \\
\rowcolor{LightCyan}
DINO & BC-SSL-T & 7$\times$7  & 46.1 & 67.4 & 50.8 & 41.2 & 64.5 & 44.2 \\
\rowcolor{LightCyan}
DINO & BC-SSL-T & 9$\times$9  & \textbf{46.3} & \textbf{67.6} & \textbf{51.0} & \textbf{41.3} & \textbf{64.7} & \textbf{44.4}  \\
\hline
  \rowcolor{Gray}
  \multicolumn{9}{c}{\em 300-epoch SSL pre-training}  \\
DINO~\citep{li2021efficient} & Swin-T & - & 46.2 & 67.9  & 50.5 & \textbf{41.7}  & 64.8 & \textbf{45.1} \\
EsViT~\citep{li2021efficient} & Swin-T & -  & 46.2 & 68.0  & 50.6 & 41.6  & 64.9 & 44.8 \\
\rowcolor{LightCyan}
DINO & BC-SSL-T & 9$\times$9  & \textbf{46.6} & \textbf{68.1} & \textbf{51.3} & 41.6 & \textbf{65.0} & 44.5 \\

\hline
\end{tabular}}
\label{tab:COCO}
\vspace{-0.5em}
\end{table*}

\paragraph{Implementation details.}
\looseness=-1 To conduct an apple-to-apple comparison between BC-SSL and Transformers,  we follow the current state-of-the-art SSL-trained Transformer results~\citep{caron2021emerging,li2021efficient,xie2021self} and adopt DINO~\citep{caron2021emerging} as our SSL framework. We construct two variants of BC-SSL, BC-SSL-T and BC-SSL-S, to be of similar sizes to Swin-T and Swin-B.  We then train them with the full 300-epoch training recipe as reported in~\citet{caron2021emerging}. Concretely, all models are trained with AdamW~\citep{loshchilov2018decoupled} and a batch size of 512, with a learning rate scaled as $lr = 0.0005*\text{batchsize} / 256$, decayed with a cosine schedule. The temperature of the teacher linearly increases from 0.04 to 0.07 within the first 30 epochs as a warm-up phase, while the temperature of the student is set to be 0.1 constantly. As suggested by~\citet{caron2021emerging}, we remove the last layer normalization of the DINO head for improving stability. Following BYOL~\citep{grill2020bootstrap} and DINO, we choose color jittering, Gaussian blur and solarization for the data augmentations and multi-crop~\citep{caron2020unsupervised} with a bicubic interpolation.

\paragraph{Evaluation protocols.} Same as~\citet{caron2021emerging,li2021efficient,xie2021self,chen2021empirical}, we report top-1 linear probe and $k$-NN accuracy on ImageNet-1K validation set. For linear probing, random size cropping and horizontal flips are adopted as augmentation. The backbone is frozen and only the classifier is trained by SGD for 100 epochs with a small learning rate of 0.001, which is decayed with a cosine decay schedule. Besides, we also evaluate the learned representation with a simple non-parametric evaluation - $k$-Nearest neighbor (k-NN) classifiers. We first store the features of the training data learned by the pre-training backbone; find the $k$ nearest features that match the feature of a test image; and vote for the label. The comparison results are reported in Table~\ref{tab:main_result_cls}. 

\paragraph{Comparisons with ResNets.} We first compare the performance of BC-SSL with the standard ResNet-50 (23M) mainly shown on the bottom panel. We observe that BC-SSL-T dramatically outperforms the best SSL-trained ResNet-50 (DINO) by 2.5\% with linear probe and 8.2\% with $k$-NN, demonstrating that the superiority of modern CNNs over the conventional ResNets can be generalized to the longer training time regime. When compared with bigger ResNets on the top panel, our 300-epoch trained BC-SSL-S with 50M parameters is able to achieve comparable performance to those large-scale ResNets with hundreds of parameters (usually trained for around 1000 epochs), while enjoying up to 5.8$\times$ higher throughput. 
  
\paragraph{Comparisons with Transformers.} We next compare BC-SSL with two strong Transformer baselines: ViT and Swin Transformer as reported on the bottom panel in Table~\ref{tab:main_result_cls}. Overall, BC-SSL provides a positive signal that modern CNNs perform satisfactorily against two strong Transformer baselines in terms of the accuracy-computation trade-off. Without using any sophisticated attention modules, BC-SSL-T outperforms Swin-T by 0.8\% with linear probing and 1.5\% with $k$-NN. Our larger model BC-SSL-S further boosts the linear classification accuracy to 79.0\%, matching the performance of Swin-S, while being slightly faster in inference throughput than the latter when tested on a V100 GPU (442 vs 437). However, when being benchmarked on the more advanced A100 GPU marked with blue colors, BC-SSL-S is 40\% faster than Swin-S (1197 vs 857), thanks to the efficient convolutional modules and simple design choices. Moreover, the promise of BC-SSL also holds when compared to the Transformers trained with contrastive methods, i.e., ViT-S trained with MoCo-v3, achieving 2.3\% higher accuracy.

\subsection{Evaluation on Downstream COCO Object Detection and Segmentation} One advantage of convolutional architectures compared with Transformers with global self-attention~\citep{vaswani2017attention,dosovitskiy2021an} is its preferable low computational complexity with respect to image size, allowing it to be efficiently transferred to downstream tasks with high resolution. To evaluate the transferability of the learned representations of BC-SSL, we also conduct transfer learning on MS COCO object detection and segmentation~\citep{lin2014microsoft}. Following previous arts~\citep{li2021efficient,xie2021self,liu2022convnet}, We finetune Mask R-CNN~\citep{he2017mask} on the COCO dataset with BC-SSL backbone for a 3$\times$ (36 epochs) schedule. Layer-wise learning rate decay~\citep{bao2021beit} and stochastic depth rate are adopted. The hyperparameters are exactly the same as the ones reported in supervised ConvNeXt~\citep{liu2022convnet}. By using this set of hyperparameters and configurations, we not only can conduct a fair comparison among various SSL results, but also can compare our self-supervised models with their supervised counterparts. To better understand the behavior of BC-SSL in different SSL training regimes, we evaluate two groups of BC-SSL models: 100-epoch pre-trained models and 300-epoch pre-trained models. Table~\ref{tab:COCO} shows the results.  We summarize the main observations below:

\textbf{\circled{1} Performance increases as the kernel size.} In the middle group of Table~\ref{tab:COCO}, we report the performance of BC-SSL (trained in 100 epochs) with increasing kernel sizes from 3$\times$3 to 9$\times$9. We can observe a very clear trend that the performance increases as the kernel size. While BC-SSL-T with 3$\times$3 kernels suffers from a big performance gap to 300-epoch pre-trained Swin-T, it gradually approaches and eventually matches the performance of the latter using 9$\times$9 kernels. The performance makes sense since larger kernels obtain a larger effective receptive field (ERF) and benefit more on the high-resolution dense prediction tasks~\citep{liu2022more}. This result highlights the better transferability of large-kernel convolutions over the Transformers on dense prediction downstream tasks.

\textbf{\circled{2} BC-SSL outperforms self-supervised Transformers.} When pre-trained with the full training recipe in 300 epochs, BC-SSL-T further boosts the performance of self-supervised CCNs on COCO over the Swin-T by a good margin, especially in terms of the box AP.

\textbf{\circled{3} BC-SSL performs better than its supervised counterpart.} Moreover, our self-supervised BC-SSL also outperforms its supervised counterpart as reported on the top panel. Given the accuracy of self-supervised BC-SSL falls short of the supervised one reported in~\citep{liu2022convnet}, this phenomenon indicates that the large kernel design in SSL brings more benefits to downstream tasks than the pre-training ImageNet task. 
\begin{table*}[h]
\small
	\caption{\textbf{Robustness evaluation of BC-SSL.} All results are obtained by directly testing our ImageNet-1K linear probing models on several robustness benchmark datasets. We do not make use of any specialized modules or additional fine-tuning procedures. }
	\label{tab:robust_BC-SSL}
  	\vspace{-0.1in}
	\begin{center}
		\resizebox{0.9\textwidth}{!}{
		\begin{tabular}{l|c|c|c|cccccc}
			\hline
                Method & Architecture & Kernel Size & FLOPs (G) / \#Param. (M) $\downarrow$ & Clean (linear) $\uparrow$ & Clean ($k$-NN) $\uparrow$ & C $\downarrow$ & SK $\uparrow$ & R $\uparrow$ & A $\uparrow$\\
                \hline
                
      \rowcolor{Gray}
      \multicolumn{10}{c}{\em 100-epoch SSL pre-training}  \\
      DINO & ResNet-50 & 3$\times$3 &  4.1 / 23 & 67.1 & 58.9 & 
      71.85 & 14.77 & 15.02 & 0.76\\
      DINO & ViT-S & - & 4.6 / 21 & 73.5 & 68.8 & 59.57 & 20.08 &21.89 &3.04 \\
      DINO & Swin-T & - & 4.5 / 28 & 75.0 &  69.7 &  61.86 & 20.16 & 20.18 & 3.49\\
      \rowcolor{LightCyan}
	  DINO & BC-SSL-T &  3$\times$3 & 4.4 / 28  & 74.0 & 68.4 &  61.82 & 23.18 & 21.60 & 3.80 \\
	\rowcolor{LightCyan}
      DINO & BC-SSL-T &  5$\times$5 & 4.4 / 28 & 74.4 & 68.8 & 61.48 & 23.45 & 22.92 & 4.75\\
      \rowcolor{LightCyan}
      DINO & BC-SSL-T &  7$\times$7 & 4.5 / 29 & 74.5 & 69.4 & 60.40  & 24.55 & 22.92 & 4.74\\
     \rowcolor{LightCyan}
      DINO & BC-SSL-T &  9$\times$9 & / 29 & \textbf{75.0} & \textbf{70.3} & 59.39 & 24.55 & 22.79 & 4.66\\
      \rowcolor{LightCyan}
      DINO & BC-SSL-T &  15$\times$15  & 4.8 / 30 & 74.5 & 69.7 & \textbf{58.80} & \textbf{25.33} & \textbf{23.25} & \textbf{5.31} \\
         \hline
          \rowcolor{Gray}
        \multicolumn{10}{c}{\em 300-epoch SSL pre-training} \\
         DINO & ResNet-50 & 3$\times$3 &  4.1 / 23 & 75.3 & 67.5 & 70.28 &14.24 &14.4  & 0.88\\
         DINO & Swin-T & - & 4.5 / 28 & 77.0 & 74.2 & 59.39 & 21.96 & 21.18 & 5.34\\
               \rowcolor{LightCyan}
         DINO & BC-SSL-T & 9$\times$9 & 4.5 / 29 & \textbf{77.8} & \textbf{75.7} & \textbf{57.44} &  \textbf{25.32} & \textbf{23.82} & \textbf{5.72} \\
           \rowcolor{LightCyan}
	\hline
		\end{tabular}}
	\end{center}
\end{table*}

\vspace{-1.5em}
\subsection{Evaluation on Robustness}


Recent studies on out-of-distribution robustness~\citep{bai2021transformers,paul2022vision,zhang2022delving,mao2022towards} show that Transformers are much more robust than CNNs when testing under distribution shifts. For instance, Mao et al.~\citep{mao2022towards} demonstrate that DeiT~\citep{touvron2021training}, Swin-T, and RVI~\citep{mao2022towards} achieve stronger robustness than ResNet. Given the advanced architecture designs in BC-SSL, we ask if the modern large-kernel CNNs can launch a successful counterattack in self-supervised learning? 

To answer this question, we directly test our linear probing classification models on several robustness benchmarks including ImageNet-A~\citep{hendrycks2021natural},
ImageNet-R~\citep{hendrycks2021many}, ImageNet-Sketch~\citep{wang2019learning} and ImageNet-C~\citep{hendrycks2019benchmarking} datasets. Mean corruption error (mCE) is reported for ImageNet-C and top-1 accuracy is reported for the rest of datasets.

We again observe an encouraging trend in Table~\ref{tab:robust_BC-SSL}, that is, the robustness of BC-SSL monotonously improves as the kernel size scales up to 15$\times$15. First, Swin-T indeed achieves better robustness performance than ResNet-50 in both settings, confirming the findings in~\citep{bai2021transformers,mao2022towards}.  It is then interesting to observe that BC-SSL with the smallest 3$\times$3 kernels is already more robust than Swin-T, and our 9$\times$9 model further performs an all-around win over Swin-T including both clean and robust accuracy, being blessed by large kernels. It is worth noting that while the 15$\times$15 kernel undergoes a small clean accuracy drop compared to 9$\times$9 kernel, it brings a notable improvement in robustness demonstrating the promise of large kernels in the context of robustness.

\begin{wraptable}{r}{0.5\textwidth}
\vspace{2 em}
 \centering
\setlength\tabcolsep{1.5pt}
\settowidth\rotheadsize{Radcliffe Cam}
\begin{tabularx}{0.5\textwidth}{l XXXX }
&\multicolumn{2}{|c|}{Grad-CAM}&\multicolumn{2}{c|}{Eigen-CAM}\\
\rothead{\centering{Original}} 
                        &   \includegraphics[width=\hsize,valign=m]{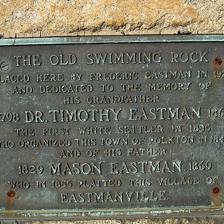}   
                        &   \includegraphics[width=\hsize,valign=m]{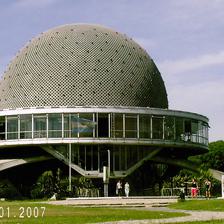}  
                        &   \includegraphics[width=\hsize,valign=m]{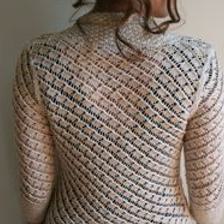}
                        &   \includegraphics[width=\hsize,valign=m]{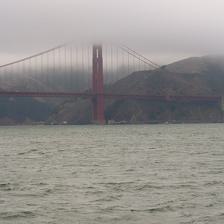}   \\[-1 em]
\rothead{\centering{RN50}} 
                        &   \includegraphics[width=\hsize,valign=m]{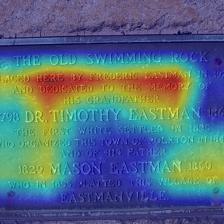}    
                        &   \includegraphics[width=\hsize,valign=m]{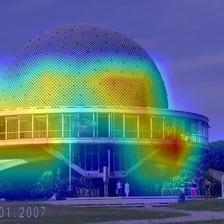}    
                        &   \includegraphics[width=\hsize,valign=m]{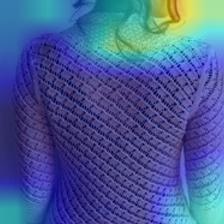}
                        &   \includegraphics[width=\hsize,valign=m]{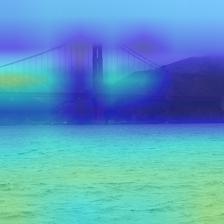}    \\[-1em]
\rothead{\centering{BC-SSL-3}} 
                        &   \includegraphics[width=\hsize,valign=m]{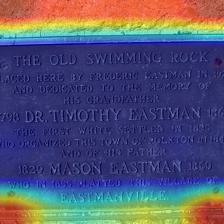}    
                        &   \includegraphics[width=\hsize,valign=m]{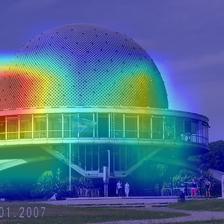}    
                        &   \includegraphics[width=\hsize,valign=m]{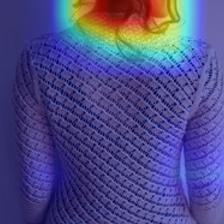}
                        &   \includegraphics[width=\hsize,valign=m]{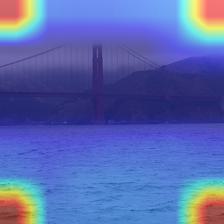}    \\[-1em]
\rothead{\centering{BC-SSL-7}} 
                        &   \includegraphics[width=\hsize,valign=m]{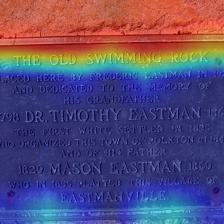}    
                        &   \includegraphics[width=\hsize,valign=m]{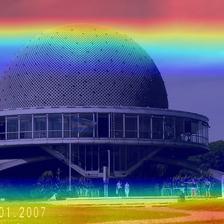}    
                        &   \includegraphics[width=\hsize,valign=m]{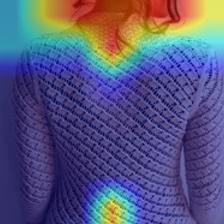}
                        &   \includegraphics[width=\hsize,valign=m]{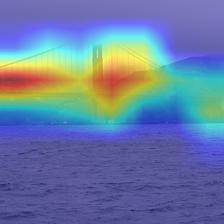}    \\[-1 em]
\rothead{\centering{BC-SSL-9}} 
                        &   \includegraphics[width=\hsize,valign=m]{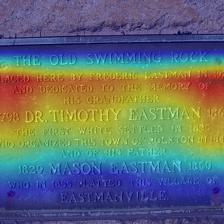}    
                        &   \includegraphics[width=\hsize,valign=m]{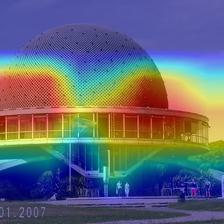}    
                        &   \includegraphics[width=\hsize,valign=m]{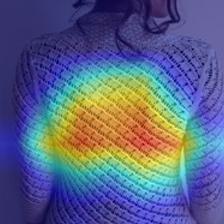}
                        &   \includegraphics[width=\hsize,valign=m]{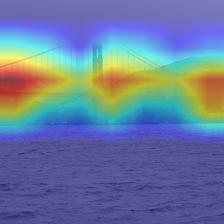}    \\[-1em]
\rothead{\centering{BC-SSL-15}} 
                        &   \includegraphics[width=\hsize,valign=m]{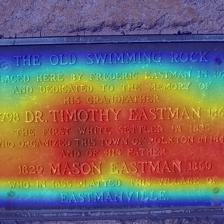}    
                        &   \includegraphics[width=\hsize,valign=m]{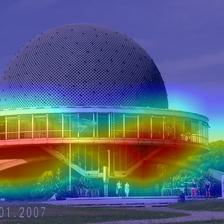}    
                        &   \includegraphics[width=\hsize,valign=m]{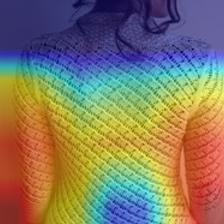}
                        &   \includegraphics[width=\hsize,valign=m]{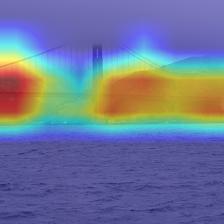}    \\[-1em]
\rothead{\centering{ViT-S}} 
                        &   \includegraphics[width=\hsize,valign=m]{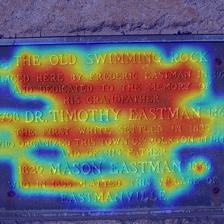}    
                        &   \includegraphics[width=\hsize,valign=m]{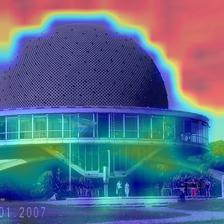}    
                        &   \includegraphics[width=\hsize,valign=m]{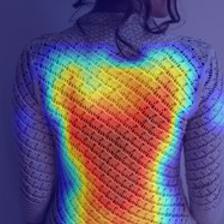}
                        &   \includegraphics[width=\hsize,valign=m]{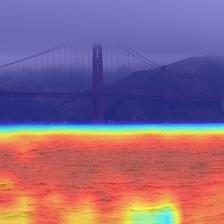}    \\[-1em]
\rothead{\centering{Swin-T}} 
                        &   \includegraphics[width=\hsize,valign=m]{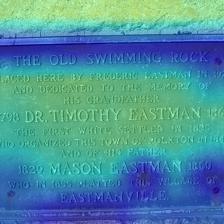}    
                        &   \includegraphics[width=\hsize,valign=m]{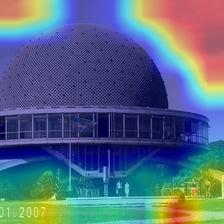}    
                        &   \includegraphics[width=\hsize,valign=m]{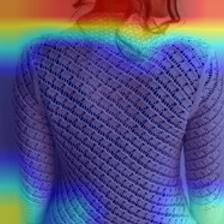}
                        &   \includegraphics[width=\hsize,valign=m]{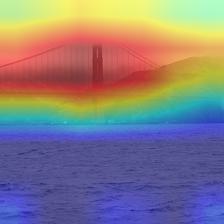}  \\ [-1em]
\end{tabularx}
\caption{Heatmap visualization generated by different models.}
\label{Fig:heatmap}
\vskip -0.3 in
\end{wraptable}

\vspace{-0.5 em}
\section{Qualitative Study}



\subsection{Visualization}

In this section, we provide visualizations through two popular tools Grad-CAM~\citep{selvaraju2017grad} and Eigen-CAM~\citep{muhammad2020eigen} to understand the mechanism discrepancy behind the decisions made by different SSL-trained architectures. 
Following the tutorials in~\citep{jacobgilpytorchcam}, we choose the representation learned by the last layer 
from the last stage to visualize CNN models. For Transformers, we choose the features after the first LayerNorm layer in the last stage's last block. Since the activation of Transformers is usually not 2D, we further reshape it to 2D spatial images. 

We compare across ResNet-50, ViT-S, Swin-T, and BC-SSL-T with various kernel sizes in Figure~\ref{Fig:heatmap}. From the heatmaps, we can conclude that CNNs with 3$\times$3 kernels (ResNet-50 and BC-SSL) either capture the smallest range of important pixels or no important pixels to make decisions. As the kernel size continuously increases, the red regions (corresponding to high scores) also gradually expand, and desirably cover the labeled object when the kernels are large enough, i.e., 9$\times$9, 15$\times$15. This indicates that large kernels inherently have a larger effective receptive field than smaller kernels, leading to more robust and accurate prediction. On the other hand, it seems that Transformers with self-attention tend to be good at capturing shapes than CNNs, although sometimes their red, shaped heatmaps are completely located in the background. This behavior is in line with the findings in supervised learning~\citep{diesendruck2003specific}, where ViT are reported to have a higher shape bias than CNNs, whereas CNNs usually tend to preserve textures rather than shapes.


\vspace{-1em}
\section{Conclusions}
This work does not propose a novel method or model but instead provides an empirical study on an incremental baseline inspired by the recent breakthroughs in self-supervised learning: rethinking self-supervised CNNs in the era of Transformers. Increasingly stronger results achieved by new self-supervised training recipes accompanied by advanced Transformers make people start to believe that Transformers or self-attention operations are inherently more suitable than CNNs in the context of SSL. In this paper, we decouple SSL from Transformers and ask whether self-attention-free architectures like CNN can deliver the same excellence with more advanced designs too. We provide an encouraging signal that we are able to build pure CNN SSL architectures that perform on par with or better than the best SSL-trained Transformers, by just scaling up convolutional kernel sizes besides several small tweaks. Our results highlight that the simple design of convolutional operations remains powerful in self-supervised learning.

\bibliography{reference}


\appendix
\section{Appendix}

\subsection{$k$-NN Monitor}
\begin{figure}[h]
\begin{center}
\includegraphics[width=0.35\textwidth]{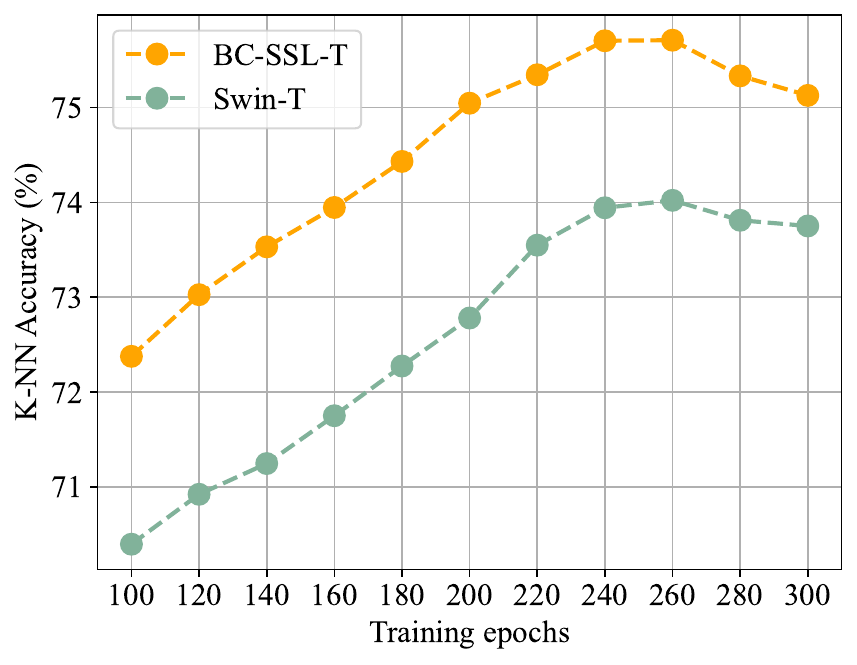}
\caption{\textbf{Training curves of different architectures using DINO.} }
\label{fig:knn_monitor}
\end{center}
\vspace{-2.2em}
\end{figure}

$k$-NN Monitor is a widely used tool to monitor training dynamics of self-supervised learning~\citep{wu2018unsupervised,chen2021empirical}. To better understand the training dynamics of different architectures, we sparsely perform $k$-NN evaluation every 20 epochs and depict the results in Figure~\ref{fig:knn_monitor}.  BC-SSL-T shares an extremely similar pattern with  Swin-T albeit with a consistently higher accuracy: an upsurging increase followed by a slight drop. The peak accuracy is reached around the 260 epoch. 

Whereas, the $k$-NN performance of ViT-S keeps improving and eventually surpasses Swin-T, closely behind BC-SSL-T. However, this promising behavior of ViT-S does not share with the linear probing scenario in Table~\ref{tab:main_result_cls}.

\end{document}